\documentclass{article}
\usepackage{PRIMEarxiv}
\usepackage[utf8]{inputenc} 
\usepackage[T1]{fontenc}    
\usepackage{hyperref}       
\usepackage{url}            
\usepackage{booktabs}       
\usepackage{amsfonts}       
\usepackage{nicefrac}       
\usepackage{microtype}      
\usepackage{lipsum}
\usepackage{graphicx}
\graphicspath{{media/}}     
\usepackage{amsmath}
\usepackage{mathrsfs}
\usepackage{multirow}
\title{\textbf{\textbf{Fast-Slow Co-advancing Optimizer: Toward Harmonious Adversarial Training of GAN}}

\thanks{\textit{\underline{Citation}}: 
\textbf{Authors. Title. Pages.... DOI:000000/11111.}} 
}

\author{
  Lin Wang$^{\dagger}$ \\
  School of Ocean Engineering \\ 
  Harbin Institute of Technology  \\
  Weihai, Shandong, China \\ 
  \texttt{wanglin\_007@hitwh.edu.cn}
  \And
  Xiancheng Wang$^{\dagger}$\thanks{$\dagger$Equal contribution} \\ 
  School of Ocean Engineering \\
  Harbin Institute of Technology\\
  Weihai, Shandong, China \\
  \texttt{402196277@qq.com}
  \And 
  Rui Wang \\
  School of Ocean Engineering \\
  Harbin Institute of Technology \\
  Weihai, Shandong, China \\
  \texttt{wangrui@hitwh.edu.cn}
  \And 
  Zhibo Zhang \\
  Technical Center, Bogie Development Department \\
  CRRC Qingdao Sifang Locomotive and Rolling Stock Co., Ltd \\
  Qingdao, Shandong, China \\
  \texttt{zhangzhibo@cqsf.com} 
  \And 
  Minghang Zhao$^{*}$\thanks{$^{*}$Corresponding author} \\
  School of Ocean Engineering \\
  Harbin Institute of Technology  \\
  Weihai, Shandong, China \\
  \texttt{zhaomh@hit.edu.cn}
}

\begin{document}
\maketitle

\begin{abstract}

Up to now, the training processes of typical Generative Adversarial Networks (GANs) are still particularly sensitive to data properties and hyperparameters, which may lead to severe oscillations, difficulties in convergence, or even failures to converge, especially when the overall variances of the training sets are large. These phenomena are often attributed to the training characteristics of such networks. Aiming at the problem, this paper develops a new intelligent optimizer, Fast-Slow Co-advancing Optimizer (FSCO), which employs reinforcement learning in the training process of GANs to make training easier. Specifically, this paper allows the training step size to be controlled by an agent to improve training stability, and makes the training process more intelligent with variable learning rates, making GANs less sensitive to step size. Experiments have been conducted on three benchmark datasets to verify the effectiveness of the developed FSCO.

\end{abstract}


\section{Introduction}
Since its inception, Generative Adversarial Networks (GANs) have remained a distinctive model in generative models, appearing in numerous fields, especially in image generation and fake data production. Its core idea of adversarial gaming, with unique training characteristics, has pioneered an era. 
\begin{figure}
    \centering
    \includegraphics[width=0.5\linewidth]{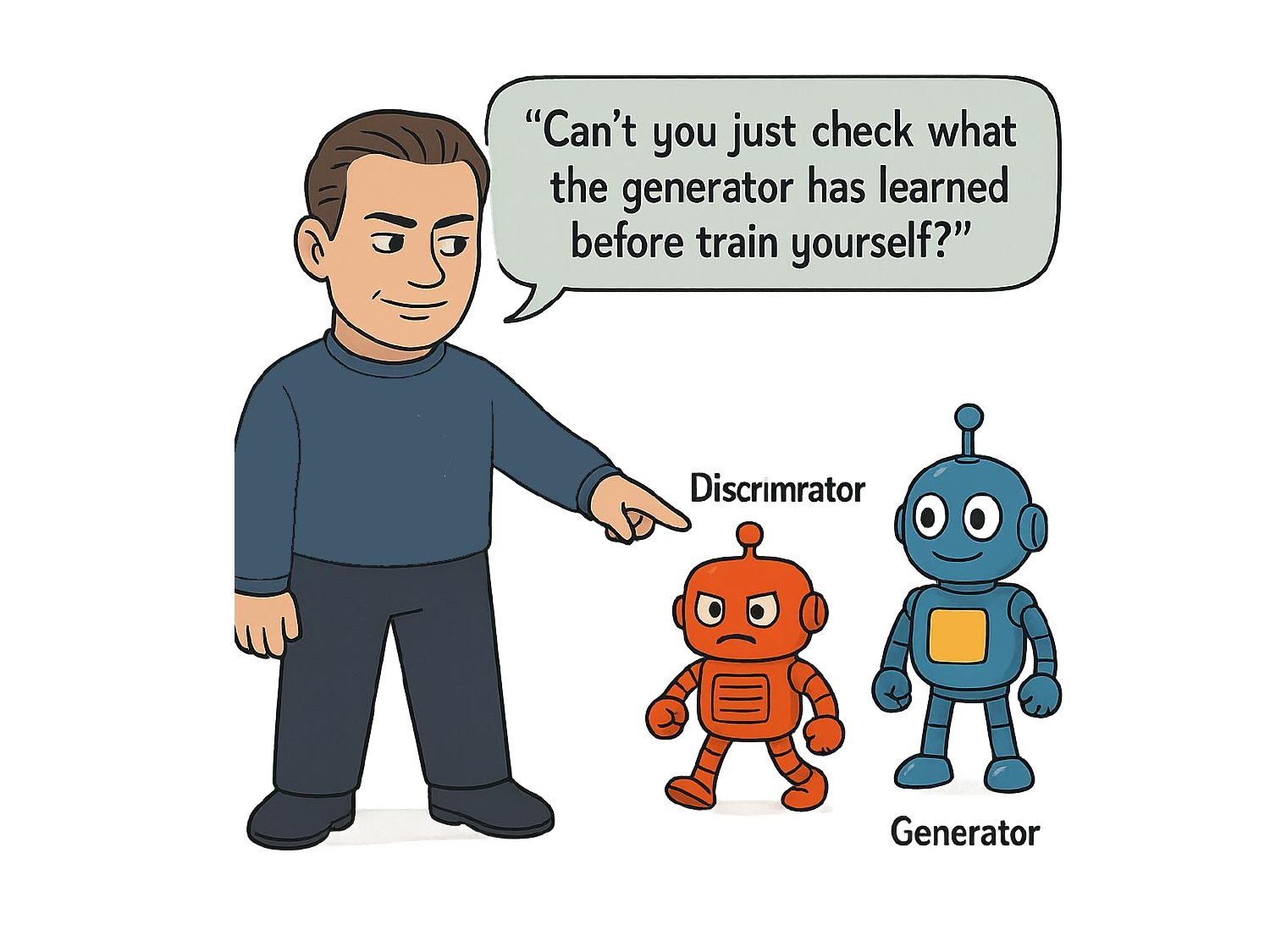}
    \caption{Dialectical thinking guided diagram}
    \label{fig:enter-label}
\end{figure}
Although the adversarial gaming idea is distinctive, its training process is often unstable. Many common phenomena, such as sudden mode collapse, gradients rapidly dropping to zero, gradient explosion, and extreme sensitivity of generators and discriminators to hyperparameters, significantly increase debugging difficulties, seriously hindering the development of GANs, and negatively affecting the mood of debugging personnel. The academic community has researched various strategies to address these difficulties. For example, Wasserstein GAN (WGAN) and WGAN with gradient penalty (WGAN-GP) adopt alternative min-max loss functions, or introduce spectral normalization regularization techniques, which indeed help alleviate instability issues in the generative adversarial processes. However, they also introduce new hyperparameters, namely penalty coefficients; these hyperparameters remain sensitive to learning rates in certain cases, increasing computational costs; and even due to the interaction between penalty terms and regularization techniques, excessive regularization may also limit model expressiveness.

To this end, this paper develops a novel training method, namely Fast-Slow Co-advancing Optimizer (FSCO), which aims to solve the problem of excessive sensitivity to step size hyperparameters in GAN training. As indicated in Figure \ref{fig:enter-label},our core idea is to let the discriminator’s learning be feedback-controlled by the loss difference between the generator and discriminator in the adversarial training process, and to let reinforcement learning agents learn how to make the discriminator wait for the generator’s learning in the adversarial game, thereby making the training process to be a relatively stable feedback control process.

Experimental results on three benchmark datasets (MNIST, ANIME, and Ganyu) show that the developed FSCO can transform the training process of GANs from an artificially tuned gaming process into a feedback-controlled training process regulated by rewards of reinforcement learning. The introduction of FSCO enlarges the hyperparameter range for coordinating step sizes of discriminator and generator, making the adversarial gaming process more harmonious. Therefore, this work provides a new approach to overcoming the training difficulties that have long plagued GAN development.

\section{Literature Review}
\label{sec:headings}

\subsection{Generative Adversarial Networks (GAN)}
The development of GANs has spanned over 10 years. Goodfellow et al.\cite{goodfellow2014generative} initiated the era of GANs in 2014. Radford et al.\cite{radford2015unsupervised} proposed the DCGAN model in 2015, replacing fully connected layers with CNN layers, which improved image generation effects and moderately reduced the difficulty of generating images. CycleGAN is another GAN variant proposed by Zhu et al. \cite{zhu2017unpaired}in 2017. Karras et al. \cite{karras2019style} introduced StyleGAN in 2019 and proposed StyleGAN2\cite{karras2020analyzing} in 2020. The entire StyleGAN series introduced style control, achieving both style manipulation and visual realism; even in 2025, an era where diffusion models shine brightly, the high quality and style control of StyleGAN2-generated images still have their unique advantages.

Since the existence of GAN, training difficulty has been a long-standing problem. To address this problem, Arjovsky et al. \cite{arjovsky2017wasserstein} improved the original GAN into WGAN in 2017, introducing Wasserstein distance to replace the traditional Jensen-Shannon divergence, changing the discriminator’s loss function to Wasserstein distance, and modifying the generator’s loss function to minimize the discriminator’s evaluation of generated samples, which partially solved the training difficulties; although WGAN brought a significant increase in training costs, it remains one of the core methods for resolving training difficulties. Gulrajani et al. \cite{gulrajani2017improved}introduced a gradient penalty mechanism based on WGAN in 2017, replacing weight clipping, alleviating the training instability caused by gradient penalties in WGAN, making the training more stable and reliable. Zhang  et al. \cite{tang2020consistency} proposed a regularization technique in 2020 to solve instability issues in GAN training. However, if the hyperparameters (especially step sizes) of discriminator and generator are inappropriate, the typical solution is repeatedly trying many times to find a suitable set of hyperparameters, and hence the training difficulty problem still exists.

\subsection{Reinforcement Learning(RL)
}
Reinforcement Learning (RL) is another branch of machine learning, representing artificial agents that achieve maximum rewards through multiple trial-and-error explorations.

According to the "Law of Effect" proposed by Thorndike \cite{thorndike1911animal} in 1911, satisfaction should be reinforced; in other words, satisfaction should be rewarded. Pavlov \cite{pavlov1927conditioned} introduced the term "reinforcement" in 1927, representing a learning methodology of stimulus and response, beginning to have theoretical and experimental foundations. Bellman \cite{bellman1957dynamic} proposed the Bellman equation in 1957, a landmark mathematical method that provided core equations for value function and value evaluation, establishing a central framework for future reinforcement learning. In 1998, Sutton et al. \cite{sutton1998reinforcement} jointly researched methods based on temporal difference and Monte Carlo, which is another milestone in the entire field of reinforcement learning. In 2013, Google’s DeepMind team proposed Deep Q-Network (DQN) \cite{mnih2013playing} by combining deep learning and Q-learning, which accomplished end-to-end learning and brought reinforcement learning into public view. Silver et al. \cite{silver2014deterministic} proposed the DPG algorithm in 2014, solving the problem of outputs not being able to be very small continuous actions. DDPG was proposed by Lillicrap et al. \cite{lillicrap2015continuous} in 2015, introducing deep learning to handle more complex states with larger state spaces. In 2015, Schulman et al. \cite{schulman2015trust} proposed the TRPO algorithm, which solved the problem of unstable policy optimization but was also overly complex. Mnih et al. \cite{mnih2016asynchronous} proposed the A3C algorithm in 2016, constructing an asynchronous parallel framework that allowed multiple agents to train in their independent environments, improving training efficiency. Schulman et al. \cite{schulman2017proximal} proposed the PPO algorithm in 2017 by improving TRPO, which remains one of the mainstream algorithms today. Fujimoto \cite{fujimoto2018addressing} et al. proposed the TD3 algorithm in 2018, solving the value function approximation problem in DDPG. Haarnoja et al. \cite{haarnoja2018soft} introduced the maximum entropy principle into policy optimization, proposing the SAC algorithm. Silver et al.\cite{schrittwieser2020mastering} proposed MuZore in 2020, enabling agent learning and planning without any knowledge of rules, and achieving a 4\% reduction in terms of bitrate in YouTube video compression by improving codecs, expanding directions for real-world applications of reinforcement learning. In 2024, large language model-guided reinforcement learning is one of the hot research areas, showing progress in multi-agent and decision systems, but still requiring exploration.

Reinforcement learning research on step size (learning rate) optimizers was first proposed by Andrychowicz et al.\cite{andrychowicz2016learning} in 2016, using RNNs, particularly LSTMs, as step size optimizers to replace traditional step size optimizers. Through meta-learning, information from the training process of the target model (e.g., historical gradients, losses) serves as input, and the trained RNN outputs the required step size. This established the paradigm of reinforcement learning as a step size optimizer and introduced the L2O (Learning to Optimize) concept. In 2017, Wichrowska et al. \cite{wichrowska2017learned}improved the LSTM network by introducing better structures and training techniques, enabling step size optimizers to better scale to large neural networks. Xu et al. \cite{xu2019learning}in 2019 proposed using RL as a step size optimizer, aiming to achieve better adaptability than traditional learning rate controllers and to obtain more appropriate learning rates. In 2023, Subramanian et al. \cite{subramanian2023meta} successfully trained agents using PPO to automatically adjust learning rates based on state changes. In 2024, Lan et al.\cite{lan2024optim4rl} proposed Optim4RL, a step size optimizer specifically designed for RL with a unique gradient processing module to address the non-stationary training problems in RL.

In summary, reinforcement learning provides an intelligent approach to hyperparameter optimization for deep learning models. However, existing reinforcement learning-based step size optimization methods typically focus on optimizing the step size for a single intelligent model and cannot be directly applied to GANs, as GANs consist of two adversarial intelligent models. Therefore, it is necessary to research reinforcement learning-based step size optimization methods specifically designed for GANs.

\section{Algorithm Introduction
}
\label{sec:others}

Since the developed FSCO integrates DDPG as an embedded module of GAN to optimize the training process, this section introduces the basics of GAN and DDPG at first, and then elaborates the details of the developed FSCO.

\subsection{Basics of GAN
}
\begin{figure}
    \centering
    \includegraphics[width=0.5\linewidth]{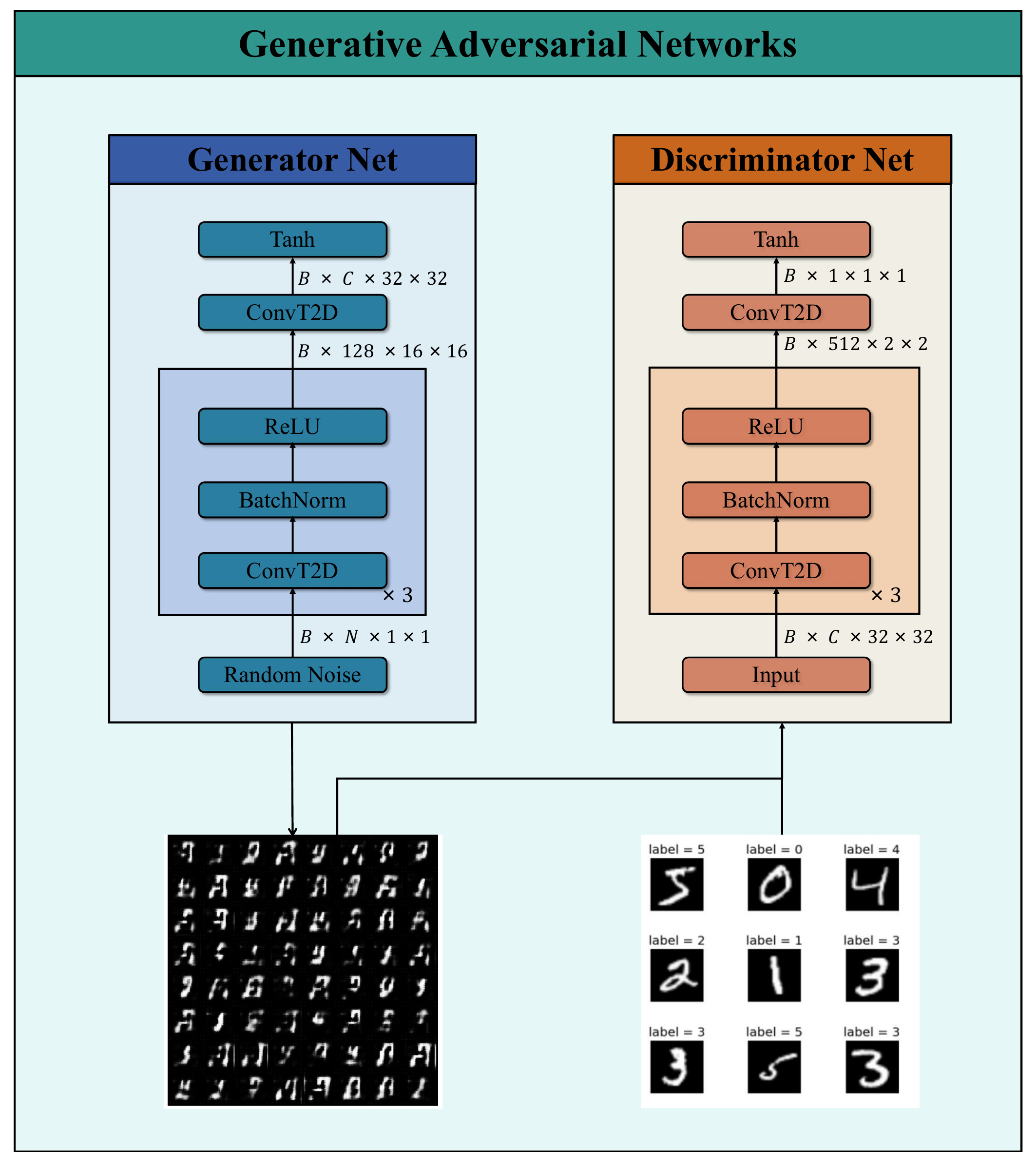}
    \caption{The Structure of GAN}
    \label{fig:enter-label}
\end{figure}
Figure 2 shows the general framework of GAN, which is composed of two neural networks: a generator and a discriminator. Through using min-max loss function, the generator aims to produce images that can deceive the discriminator with maximum probability, while the discriminator attempts to identify the generator’s fake images with maximum probability. The training process involves the procedure that the fake data produced by the generator, along with the real data, are input to the discriminator. The generator produces fake data, feeds them into the discriminator, and uses discriminator’s feedback to guide the generator’s training.

The generator and discriminator are updated through backpropagation based on the loss function in an adversarial game. In the min-max loss function, the generator and discriminator have opposing update directions. If one side becomes too powerful during the game, it cannot effectively guide the other side, so this paper aims for a harmonious process.



\subsection{Basics of DDPG
}
 \begin{figure}
     \centering
     \includegraphics[width=0.75\linewidth]{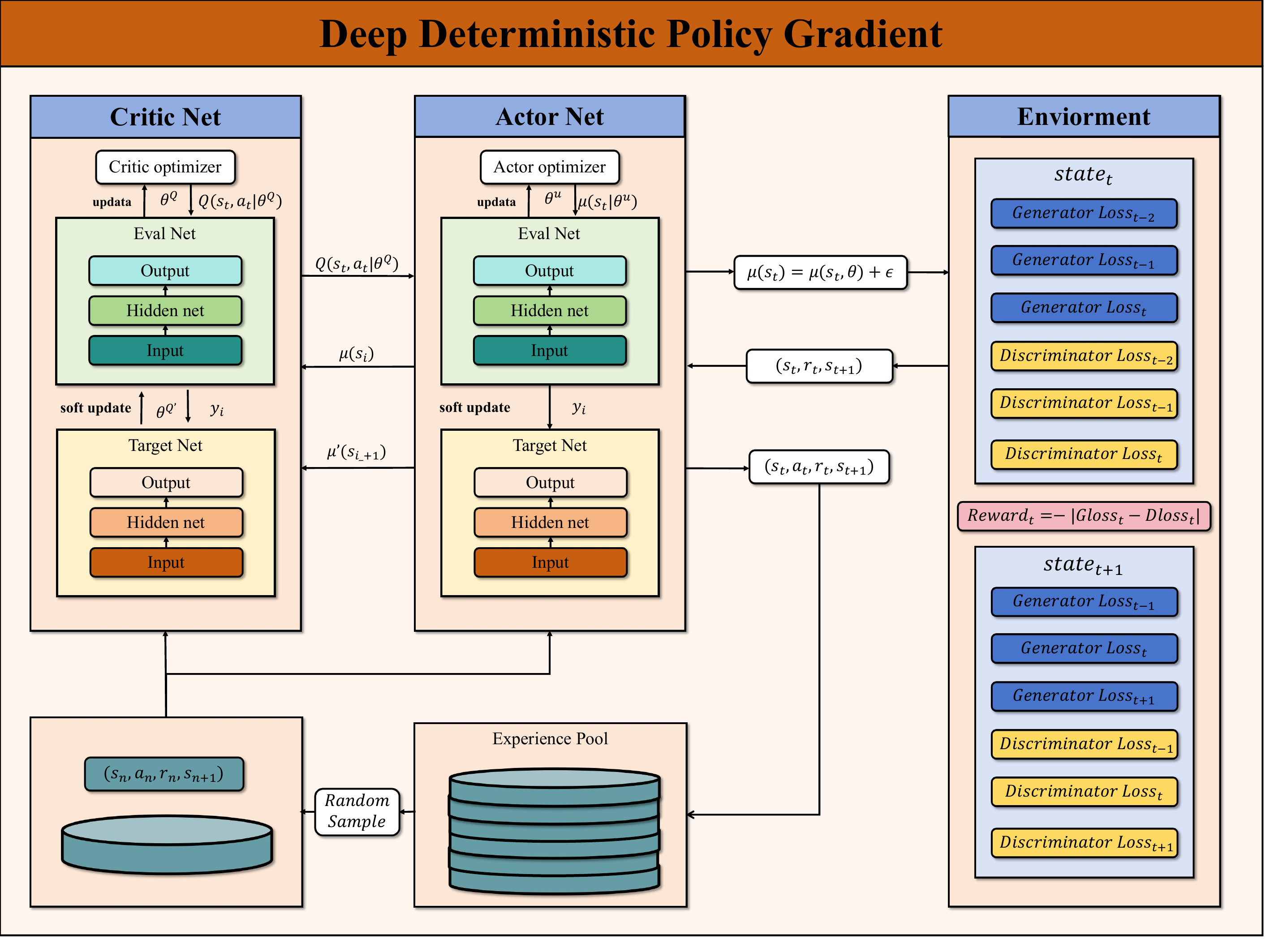}
     \caption{The Structure of DDPG}
     \label{fig:enter-label}
 \end{figure}
DDPG is one of the mainstream reinforcement learning algorithms. Its methodology is that through considering the faced states, the agent takes a series of coherent actions, and through multiple trials and iterative optimizations, ultimately maximizing the return.

As illustrated in Figure 3, the DDPG algorithm consists of two main networks: the Critic network and the Actor network. The Critic network predicts future cumulative discounted returns, while the Actor network determines what action to take in the current situation to maximize future rewards. Through repeated interactions between the agent and the environment, this method iteratively updates the network, ultimately forming an agent that outputs actions adapted to the training scenario and maximizes expected rewards. DDPG also relies on an experience replay pool, as shown in Figure 3, which uses random sampling of stored data for DDPG training, making the training process more stable. It also has a soft update mechanism, as depicted in Figure 3, where Critic-net and Actor-net have target-net and Eval-net, respectively. This is because without the soft update mechanism, directly copying Eval-net parameters to Target-net would cause severe oscillations in values, making training very unstable. The soft update mechanism is illustrated in Equation.\ref{eq:soft_update}, where  $\tau$ is typically a very small number used for the network’s soft update parameter, $\theta$  is the Eval-net parameter, $\theta'$  is the Target-net parameter.

\begin{equation}
    \theta' = \tau \theta + (1-\tau) \theta'
    \label{eq:soft_update}
\end{equation}
Through Equation \ref{eq:soft_update}, it can be found that through using  $\tau$ , each update only transfers a small portion of Eval-net parameters to Target-net, preventing update oscillations.

\subsection{The developed FSCO
}
For the training process of GANs, the step sizes of discriminators and generators are typically very important hyperparameters that need to be carefully adjusted during actual training processes, especially when facing datasets with different properties. Here, the purpose is to balance the adversarial interactions between discriminators and generators, avoiding the mode collapse problem.

\subsubsection{Automatic Setting of Discriminator Training Step Size
}
For the training process of GANs, the step sizes of discriminators and generators are typically very important hyperparameters that need to be carefully adjusted during actual training processes, especially when facing datasets with different properties. Here, the purpose is to balance the adversarial interactions between discriminators and generators, avoiding the mode collapse problem.

Through integrating DDPG to control step sizes of the discriminator, this gaming process is expected to become more harmonious. Here, step sizes directly determine the learning effects of generator and discriminator; along with gradient directions, they are two important hyperparameters. As shown in the middle of Figure 4, if the discriminator and generator game with fixed step sizes, imbalance often cause the discriminator to become too strong, preventing the generator from continuing to learn, leading to mode collapse. In contrast, if step sizes of the discriminator are properly controlled by DDPG, as shown on the right side of Figure 4, this allows the discriminator to observe the generator's learning effect and wait for the generator. Consequently, the gaming process becomes more harmonious, preventing the discriminator from learning beyond limits, thereby avoiding the mode collapse issue.

Here is the method for controlling the step size . The action output  $u(t)$ of DDPG is a number from 0 to 1. Through multiplying a fixed discriminator step size  ${\eta}_{D}$  with u(t), the ${\eta}_{FSCO-D}$  can vary within the range of [0,  ${\eta}_{D}$]. This is fundamentally controlled by DDPG’s action, and the action output is related to reward settings, which are discussed in the next section.

\begin{equation}
     {\eta}_{FSCO-D}(t)={\eta}_{D}(t)\times u(t)
     \label{eq:FSCO-Dlr}
\end{equation}

 \subsubsection{DDPG Reward Setting
}
The reward is specially designed to set proper step sizes for the discriminator. As described in Section 3.3.1, if the discriminator can observe the generator’s learning effect and intelligently adjust its step size, it can greatly control the intensity of the gaming process, making GAN be trained in a more harmonious manner, avoiding the training collapse. This paper aims to achieve the harmonious gaming like the spiral ascent process shown in the middle of Figure 5. 

\begin{figure}
    \centering
    \includegraphics[width=1\linewidth]{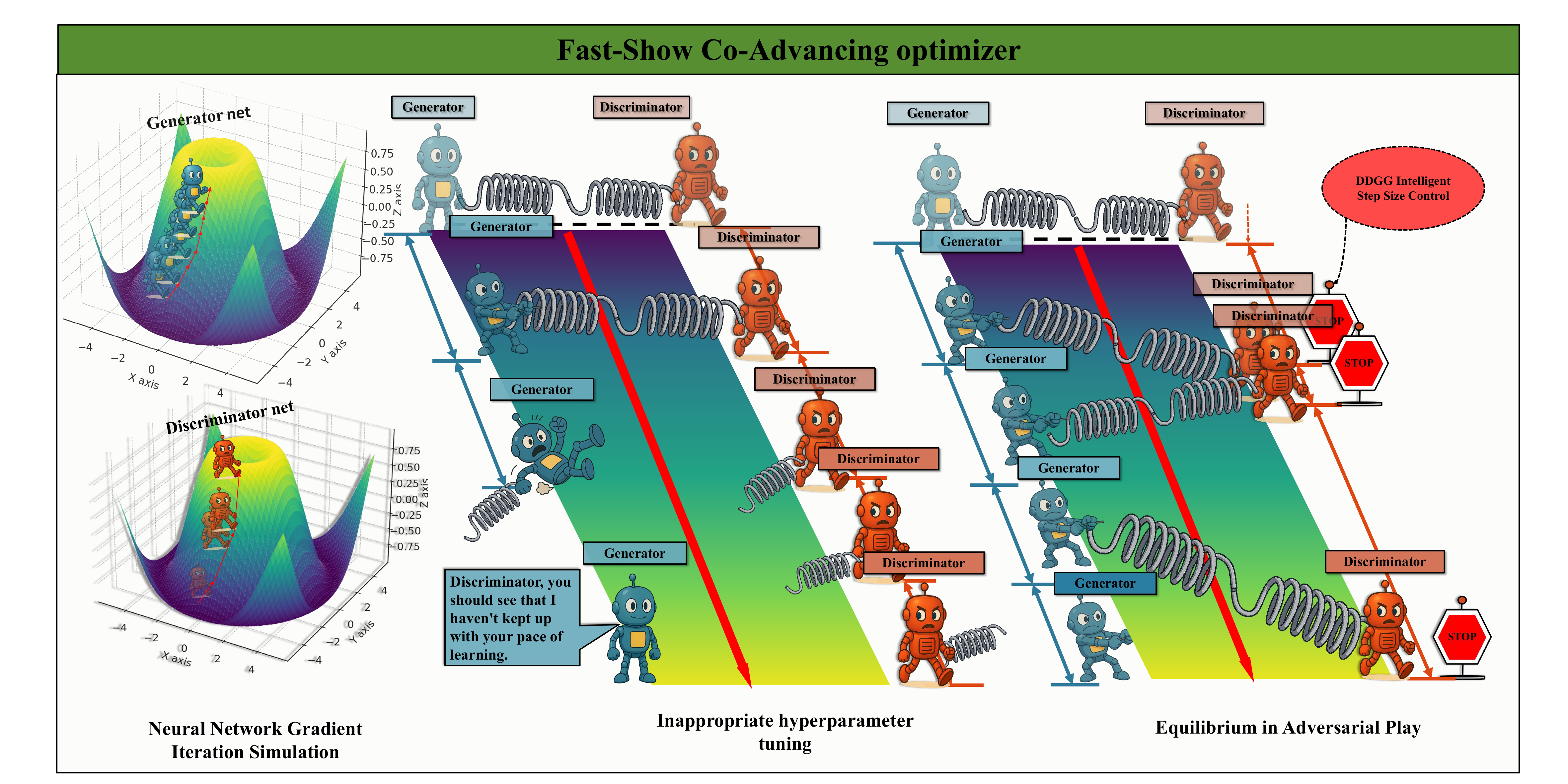}
    \caption{Harmonious adversarial training diagram in terms of step size }
    \label{fig:enter-label}
\end{figure}
As depicted in the right side of Figure 3, the DDPG reward encourages closer losses of the discriminator and the generator through using the min-max loss function.

\begin{equation}
    Reward(t) = -\left | Gloss(t) - Dloss(t))\right |
    \label{eq:rewardt}
\end{equation}

This reward Equation in   \eqref{eq:rewardt}  serves two purposes: first, if the discriminator’s loss(Dloss) is large, it can increase discriminator’s step size to accelerate discriminator learning; second, if the generator’s loss(Gloss) is too large, it can reduce discriminator’s step size to give the generator sufficient steps for learning.

In the FSCO-based training process of GAN, the phenomenon like the middle part of Figure 5 emerges, which is a progressive and harmonious gaming process that makes training more stable and the discriminator’s learning more controllable. This gaming process is described in Section 2.3.3, and the experimental results in Section 4 also provide evidence.

 \subsubsection{\textbf{Overall Steps}
}
The methodology behind the FSCO-based training of GANs (including DCGAN) is to make the training more harmonious, with a higher probability of completing the process in which the discriminator guides the generator’s learning during the adversarial interaction. As depicted in the top part of Figure 5, the hollow cube indicates DDPG, the red sphere indicates the discriminator, and the blue sphere indicates the generator. If the adversarial interaction proceeds smoothly, it will exhibit the harmonious adversarial game process shown in the middle part of Figure 5, with the specific process depicted in the right part of the figure.

During the game process, if the generator cannot improve its ability under fixed step sizes due to the discriminator’s quick learning, DDPG (represented by the black frame getting darker) suppresses the discriminator’s learning (although it still maintains a very small learning rate), allowing the generator to gradually improve its generation ability. Even though this process may not proceed as smoothly as shown in the figure, DDPG significantly inhibits the discriminator’s learning, giving the generator sufficient learning time to complete this learning process. At least, this is how the ideal harmonious process works. Besides, the gray arrow indicates the ability gap between the generator and discriminator; when the gap has not reached a sufficiently small level, DDPG maintains a small step size in long-term experience replay to inhibit the discriminator’s learning.

For ease of understanding, some explanations are provided in Figure 5. The size (radius) of the spheres represents the actual abilities of the generator and discriminator; the depth of color of the black frame represents DDPG’s suppression effect on the discriminator’s ability improvement through step size control. These descriptions are illustrated in the upper but not uppermost part of the left portion of Figure 5. The lower part of the left portion shows the procedures of DDPG in each adversarial game cycle: the parameters from the current game environment (see left part of Figure 5) are calculated to provide what DDPG needs, and DDPG gives an output  $u(t)\in\left(0,1\right)$. Afterwards, this paper uses the formula  ${\eta}_{FSCO-D}(t)={\eta}_{D}(t) \times u(t)$  (where ${\eta}_{D}(t)$ represents the fixed step size set for the discriminator), which gives the intelligently controlled discriminator step size ${\eta}_{FSCO-D}$ .  ${\eta}_{FSCO-D}$ is then input into the backpropagation process along with  ${\eta}_{G}(t)$  (the fixed step size of the generator), completing the parameter update and cycle.

The procedures of the developed FSCO are described as follows, which can be understood in conjunction with the box in the left side of Figure 5:

1. After each round of DCGAN training, the environment provides reward and state to DDPG, and then DDPG decides its output $u(t)\in\left(0,1\right)$.

2. The output is multiplied with the discriminator’s fixed step size ${\eta}_{D}(t)$ to obtain the dynamic step size ${\eta}_{FSCO-D}$.

3. ${\eta}_{FSCO-D}$ and  ${\eta}_{G}(t)$ are set as the two step sizes of discriminator and generator for the next update.

4. Perform backpropagation of gradients for discriminator and generator, and update the trainable parameters. The backward propagation of the generator is shown in Equations \ref{eq:G_wupdata}and\ref{eq:G_bupdata}. The backward propagation formula of the discriminator is shown in Equations \ref{eq:D_wupdata} and\ref{eq:D_bupdata}. $W^{(l)}$ represents the weights in layer $l$.$b^{(l)}$ represents the biases in layer $l$. Besides $V_{w^{(l)}}L$ and $V_{b^{(l)}}L$ represent the gradients of the loss function $L$ with respect to the weights and biases of layer $l$,respectively.
\begin{equation}
    W^{(l)} \leftarrow W^{(l)} - {\eta}_{G}\times V_{W^{(l)}}L
    \label{eq:G_wupdata}
\end{equation}
\begin{equation}
    b^{(l)} \leftarrow b^{(l)} - {\eta}_{G}\times V_{b^{(l)}}L
    \label{eq:G_bupdata}
\end{equation}
\begin{equation}
    W^{(l)} \leftarrow W^{(l)} -  {\eta}_{FSCO-D}(t)\times V_{W^{(l)}}L
    \label{eq:D_wupdata}
\end{equation}
\begin{equation}
    b^{(l)} \leftarrow b^{(l)} -  {\eta}_{FSCO-D}(t)\times V_{b^{(l)}}L
    \label{eq:D_bupdata}
\end{equation}

5. Complete the cycle and repeat another cycle until a predefined cycle number.

\begin{figure}
    \centering
    \includegraphics[width=1\linewidth]{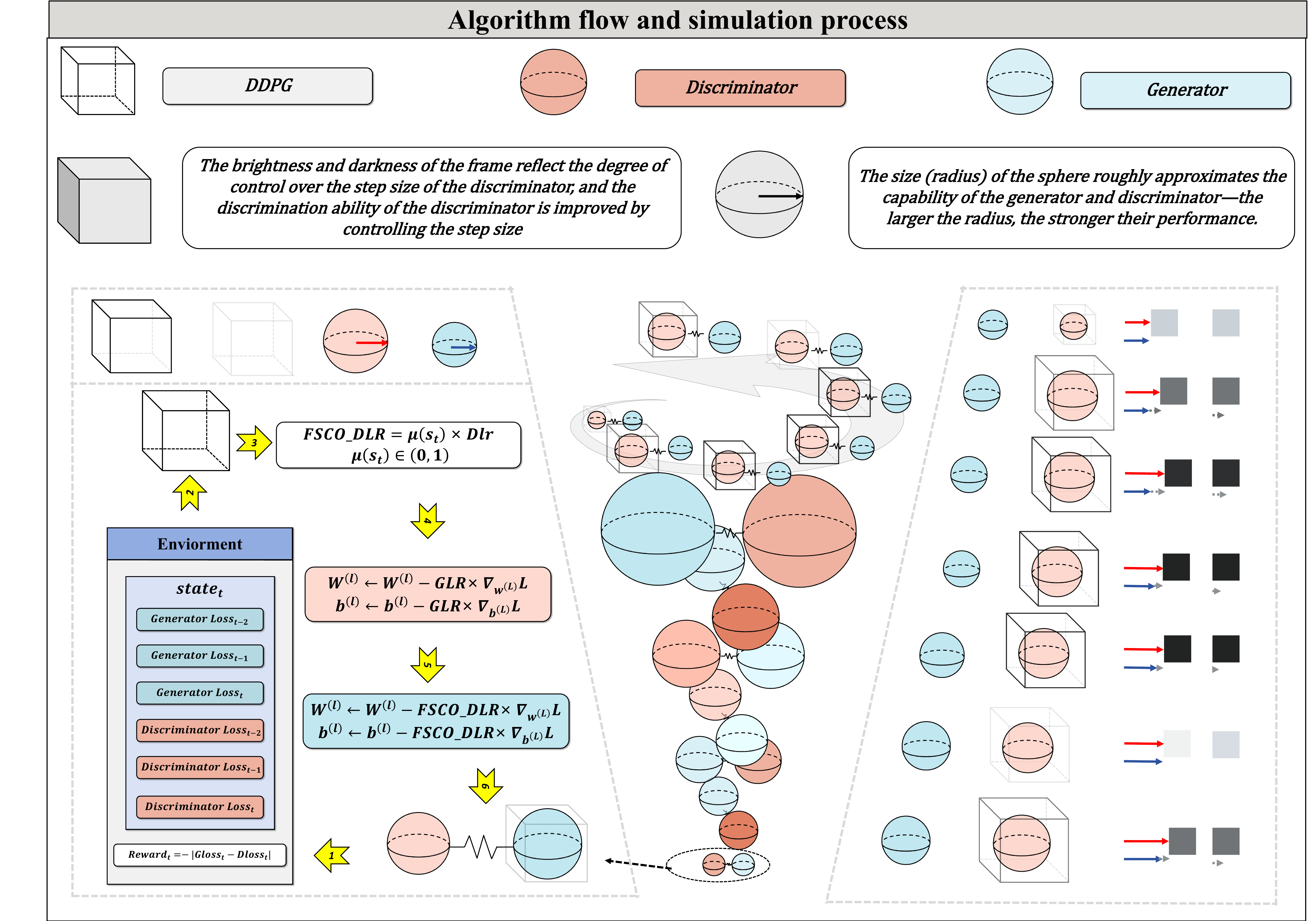}
    \caption{Algorithm diagram of the developed FSCO method }
    \label{fig:enter-label}
\end{figure}
\section{Experiments
}
\subsection{Experiment on 28×28 Image Dataset
}
\subsubsection{Hyperparameter Setup
}
An experiment has been conducted on the MNIST dataset with appropriate hyperparameters for 28×28 images. Hyperparameter Setting  is in Table \ref{tab:hyperparameters}.
\begin{table}[htbp]
\centering
\caption{Hyperparameter setup on the 28×28 image dataset}
\label{tab:hyperparameters}
\begin{tabular}{lll}
\toprule
\textbf{Methods} & \textbf{Hyperparameter} & \textbf{Value} \\
\midrule
\multirow{7}{*}{GAN Network} & Image size & 28 \\
 & Image channel number & 1 \\
 & Noise dimension & 100 \\
 & Training epochs & 50 \\
 & Generator learning rate & 0.0002 \\
 & Discriminator base learning rate & 0.002 \\
 & Batch size & 128 \\
\midrule
\multirow{9}{*}{DDPG} & State dimension & 6 \\
 & Action dimension & 1 \\
 & ACTOR LR & 0.0001 \\
 & Critic LR & 0.0001 \\
 & Discount factor & 0.99 \\
 & Target network soft update parameter & 0.005 \\
 & Experience replay buffer size & 10000 \\
 & DDPG batch size & 64 \\
 & Exploration noise & 0.1 \\
\bottomrule
\end{tabular}
\end{table}

\subsubsection{Experimental Results}

\begin{figure}
    \centering
    \includegraphics[width=0.5\linewidth]{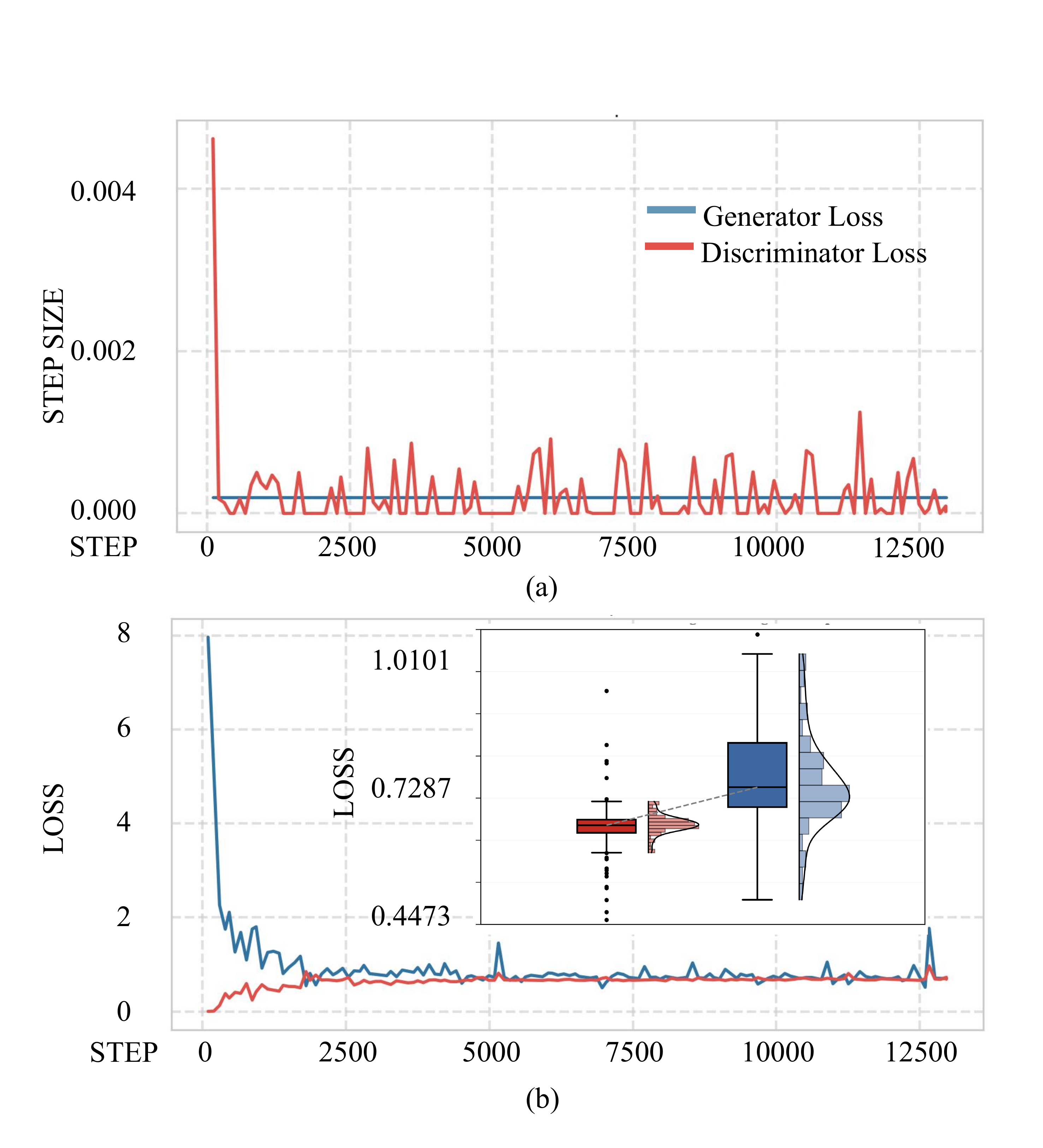}
    \caption{Step sizes and losses along the training process on the MNIST }
    \label{fig:1_1}
\end{figure}

\begin{figure}
    \centering
    \includegraphics[width=0.5\linewidth]{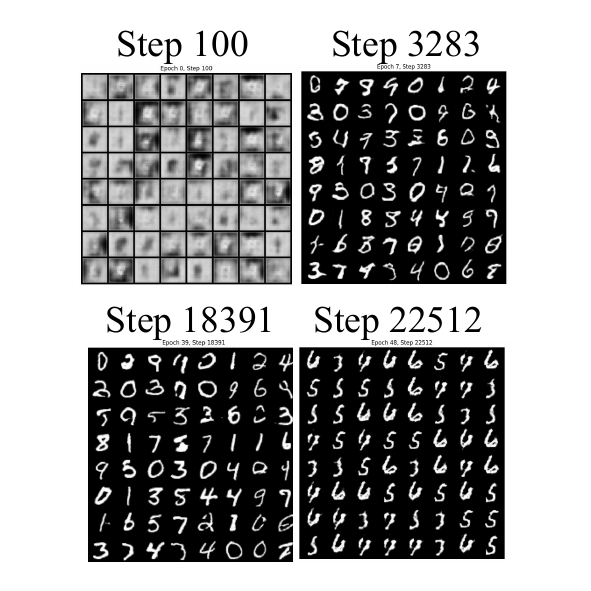}
    \caption{Generated images in size of 28$\times$28}
    \label{fig:1_2}
\end{figure}

Through applying the developed FSCO on the 28×28 public dataset, it is found that the harmonious adversarial game characteristic of this method are significant, as shown in Figure \ref{fig:1_1}. The discriminator’s learning rate becomes almost zero when the generator’s loss is slightly higher, especially after stabilizing the game between the two. Even when certain fluctuations occur, the developed FSCO can restore the process to the ideal game process within a certain range. Meanwhile, overfitting phenomena appeared in the later stages, as shown in Figure \ref{fig:1_2}, where there is obvious overfitting at step=22512.


\subsection{Experiments on 64×64 Image Dataset
}
\subsubsection{Hyperparameter Setup
}
An experiment has been conducted on the public ANIME dataset with appropriate hyperparameters for 64×64 images
.Hyperparameter Setting  is in Table \ref{tab:hyperparameters2} .

\begin{table}[htbp]
\centering
\caption{Hyperparameter setup on the 64×64 image dataset}
\label{tab:hyperparameters2}
\begin{tabular}{lll}
\toprule
\textbf{Methods} & \textbf{Hyperparameter} & \textbf{Value} \\
\midrule
\multirow{9}{*}{GAN Network} & Image size & 64×64 \\
 & Noise dimension & 100 \\
 & Training epochs & 50 \\
 & Generator learning rate & 0.0002 \\
 & Base discriminator learning rate & 0.005 \\
 & Batch size & 64 \\
 & DDPG update steps & 1 \\
 & Generator feature number & 64 \\
 & Discriminator feature number & 64 \\
\midrule
\multirow{10}{*}{DDPG} & State dimension & 6 \\
 & Action dimension & 1 \\
 & Actor Learning Rate & 0.0001 \\
 & Critic Learning Rate & 0.0001 \\
 & Discount factor & 0.99 \\
 & Target network soft update rate & 0.005 \\
 & Experience replay buffer size & 100000 \\
 & DDPG batch size & 64 \\
 & Exploration noise & 0.1 \\
 & Discriminator LR minimum multiplier & 0.001 \\
\bottomrule
\end{tabular}
\end{table}

\subsubsection{Experimental Results}
\begin{figure}
    \centering
    \includegraphics[width=0.5\linewidth]{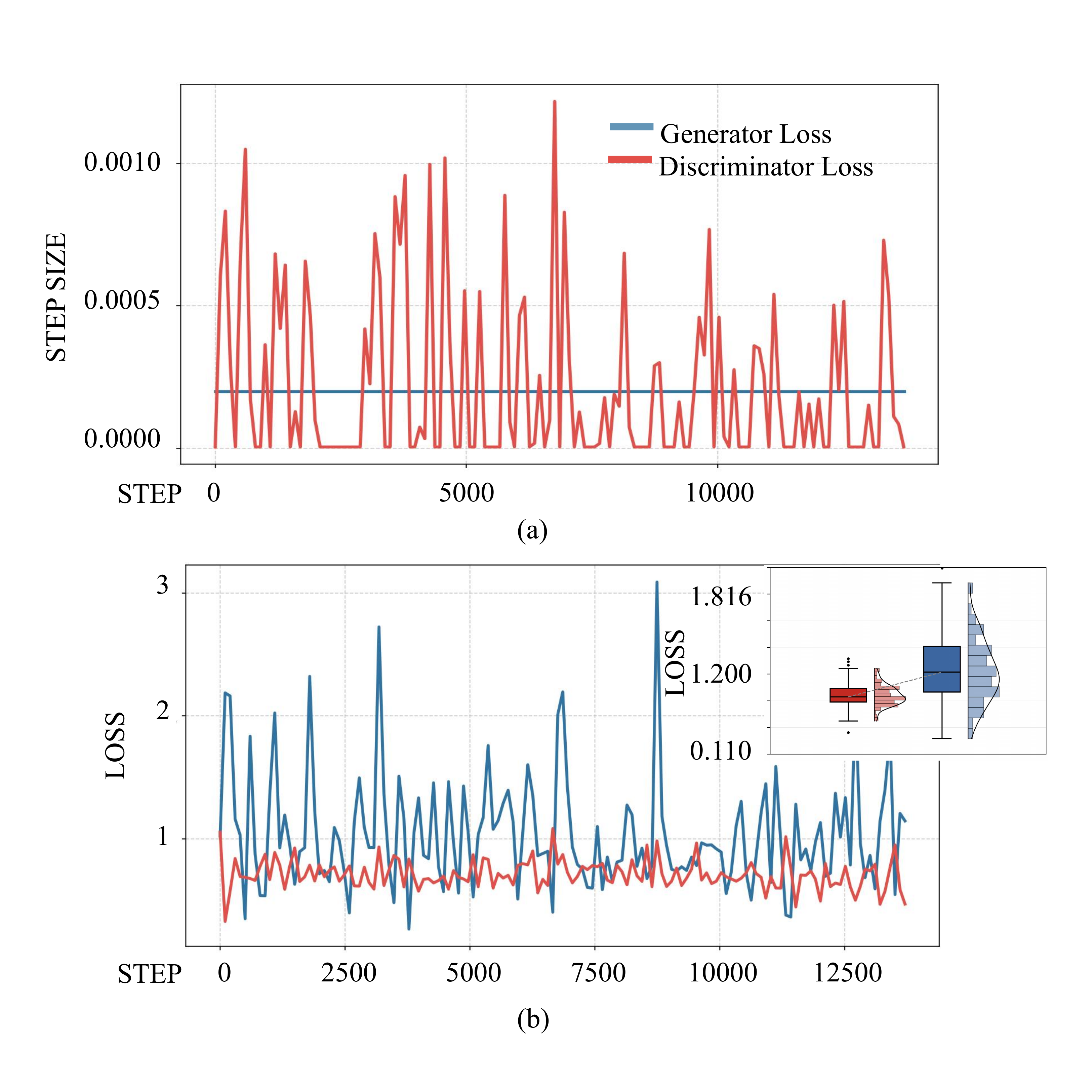}
    \caption{Step sizes and losses along the training process on the Anime}
    \label{fig:2_1}
\end{figure}

\begin{figure}
    \centering
    \includegraphics[width=0.5\linewidth]{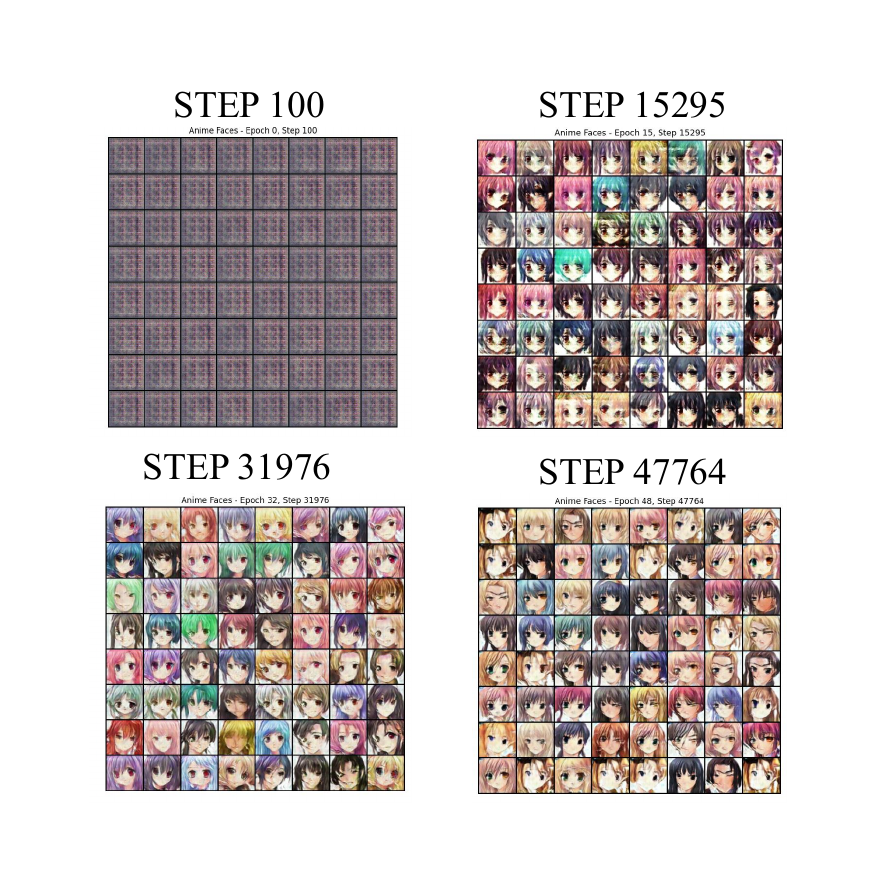}
    \caption{Generated images in size of 64$\times$64 }
    \label{fig:2_2}
\end{figure}

Although the gaming process was exceptionally intense when dealing with 64×64 images as shown in Figure \ref{fig:2_1}, the generation results in Figure \ref{fig:2_2} are still good. In other words, even in this relatively intense gaming process, FSCO was still able to produce images of reasonable quality. The harmonious gaming process brought by FSCO gave the generator sufficient time for gradient updates, ensuring the training process continued effectively, though some overfitting phenomena still occurred.


\subsection{Experiments
 on 128×128 Dataset Training}

\subsubsection{Hyperparameter  Parameter Settings}
Likewise, an experiment has been conducted on the public Ganyu dataset with appropriate hyperparameters for 128×128 images.
.Hyperparameter Setting is in Table \ref{tab:Hyperparameter3}.

\begin{table}[htbp]
\centering
\caption{Hyperparameter setup on the 128×128 image dataset}
\label{tab:Hyperparameter3}
\begin{tabular}{lcl}
\toprule
\textbf{Methods} & \textbf{Hyperparameter} & \textbf{Value} \\
\midrule
\multirow{9}{*}{GAN} & Image size & 128×128 \\
 & Image channel number & 3 \\
 & Generator input noise dimension & 100 \\
 & Training rounds & 3500 \\
 & Generator learning rate & 0.0002 \\
 & Discriminator base learning rate & 0.005 \\
 & Batch size & 64 \\
 & Generator base feature map size & 64 \\
 & Discriminator base feature map size & 64 \\
\midrule
\multirow{10}{*}{DDPG} & State dimension & 6 \\
 & Output action dimension & 1 \\
 & Actor learning rate & 0.0001 \\
 & Critic learning rate & 0.0001 \\
 & Discount factor & 0.99 \\
 & Target network update rate & 0.005 \\
 & Experience replay buffer & 100,000 \\
 & Training batch & 64 \\
 & Noise ratio & 0.1 \\
 & Discriminator minimum LR multiplier & 0.001 \\
\bottomrule
\end{tabular}
\end{table}

\subsubsection{Experimental Results}
\begin{figure}
    \centering
    \includegraphics[width=0.5\linewidth]{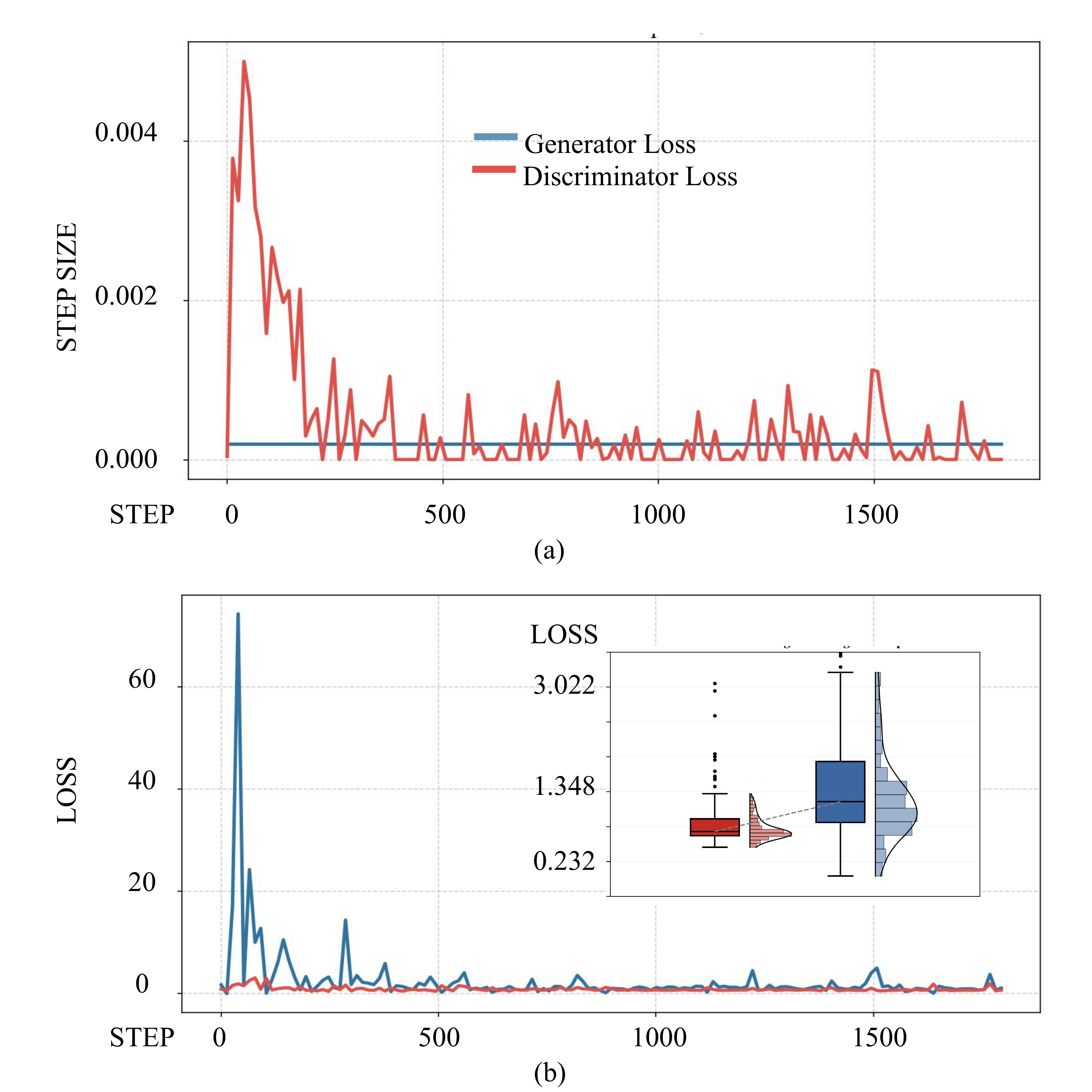}
    \caption{Step sizes and losses along the training process on the Ganyu}
    \label{fig:3_1}
\end{figure}

\begin{figure}
    \centering
    \includegraphics[width=0.5\linewidth]{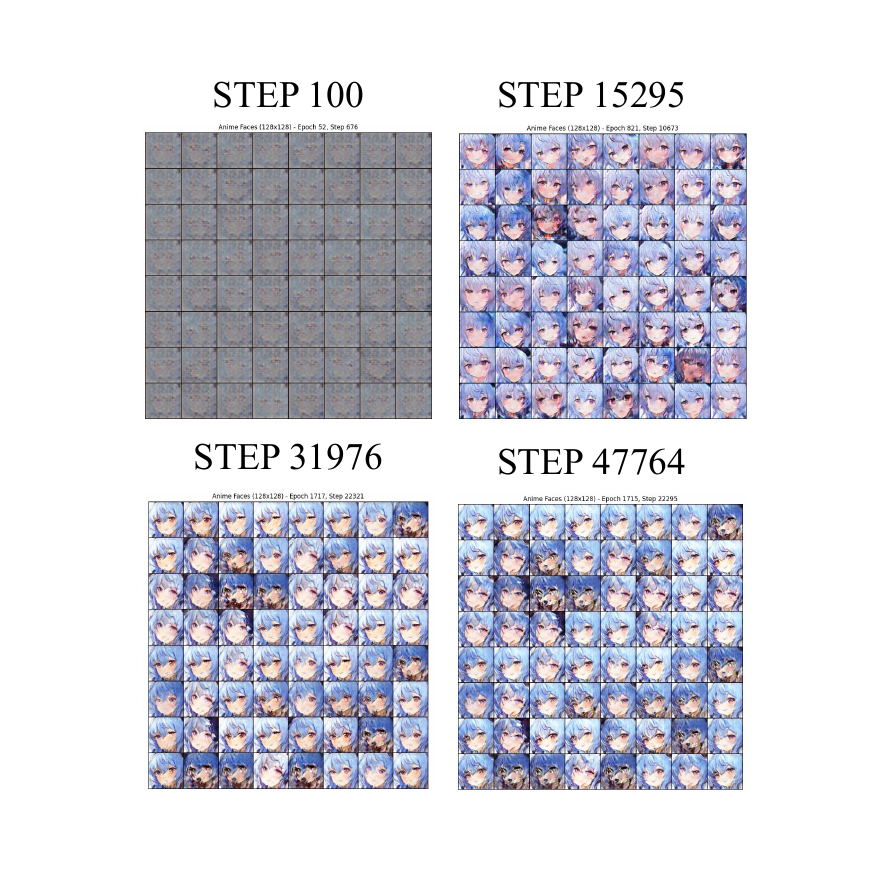}
    \caption{Generated images in size of 128$\times$128}
    \label{fig:3_2}
\end{figure}

For the 128×128 network training, the gaming process was even more intense, with the discriminator occasionally halting training to wait for the generator to learn, as shown in Figure \ref{fig:3_1}. The generator's results, shown in Figure \ref{fig:3_2}, still demonstrated obvious overfitting.

\subsection{Experimental Discussion
}

The experiments conducted were far more than just the above presented, but the differences among these experiments were mostly small. To be specific, through adjusting the discriminator’s fixed learning rate, different situations emerged, in which their differences remained relatively small and most experiments completed successfully. The harmonious gaming phenomena were observed, though with varying levels of intensity in the gaming process. Regarding whether too many iterations could lead to overfitting, the experimental results showed this was indeed the case and quite severe.

Through conducting multiple experiments, it is found that some issues still remain. First, the quality of generated images is not particularly stable - while most are of acceptable quality, there remains a small portion of generated images of inferior quality. Second, there are some uncertain with the generator’s effectiveness in the later training stages - the gaming process in late training becomes exceedingly lengthy, and it’s difficult to find out whether the generator is still learning by examining the generated images. These issues bring uncertainties to the experimental results.

Furthermore, if training results are unsatisfactory, the fixed step size (of learning rate) of discriminator should be carefully examined. The discriminator’s fixed step size has a certain theoretical basis. In theory, the step size in the FSCO method should be mostly distributed in the range
 [0, ${\eta}_{FSCO-D}(t)$], but in practice, the step size when using the FSCO has a higher probability of being zero or distributed in the 
 [0.05${\eta}_{FSCO-D}(t)$, 0.9${\eta}_{FSCO-D}(t)$]interval, with values outside this range having lower probability.Therefore, the choice of the discriminator’s fixed step size will have a certain impact on this FSCO algorithm, requiring caution when selecting it.

Typically, if the network is relatively small, the learning rate of the generator should be close to that of the discriminator, probably with the discriminator’s learning rate slightly larger than the generator’s. The above hyperparameter setups are of reference value in future applications, which also depend on the characteristics of training set. That is to way, since there is still no theoretically optimal fixed step size, the choice of the discriminator’s fixed step size still requires careful consideration. While FSCO will expand the of discriminator’s step size, the chosen step size should not be too extreme.

One noteworthy point is that FSCO cannot enhance the network’s inherent capabilities. Experiments were conducted on a 512×512 network, but the algorithm was unsuccessful – in our experiment, the discriminator was almost unable to learn features from the training set, preventing further guidance for the generator’s learning and resulting in learning failure.

Despite these existing issues, the FSCO method can still reduce the time for hyperparameter debugging. Currently, the popularity of GANs may have been surpassed by diffusion models, but they still have good application prospects in various fields. In summary, the FSCO method provides a new training approach for GANs, improving their ease of use.






\section{Conclusion}
This paper proposes a new intelligent optimizer, namely FSCO, for harmonious adversarial training of generators and discriminators of GANs (including DCGANs). By introducing DDPG to control the training step lengths of DCGAN, rather than directly using empirical adaptive optimizers or fixed step lengths, the developed FSCO is anticipated to have stronger ability in suppressing oscillations and maintaining stability during the training process, without much reliance on manual hyperparameter tuning. Even if FSCO occasionally fails, it can be easily remedied by adjusting the learning rate of discriminator. Through experiments on public datasets, this paper verifies its effectiveness in aspects of oscillation control and maintaining stable training.

\section*{Acknowledgments}
This work is supported by the National Key Research and Development Program of China (2023YFB3308100), the China State Railway Group Co., Ltd. Science and Technology Research and Development Program Project (K2024J011) and the Natural Science Foundation of Shandong Province (ZR2023ME124).

\makeatletter
\renewcommand\@biblabel[1]{[#1]}
\makeatother


\end{document}